%% file: iclr2026_conference.tex
\documentclass{article}
\usepackage{iclr2026_conference,times}
\input{math_commands.tex}

\usepackage{hyperref}
\usepackage{url}
\usepackage{booktabs}
\usepackage{multirow}
\usepackage[table]{xcolor}
\usepackage{graphicx}
\usepackage{subcaption}
\usepackage{enumitem}
\usepackage{listings}
\usepackage{tikz}

\lstdefinestyle{pycode}{
  language=Python,
  basicstyle=\ttfamily\scriptsize,
  keywordstyle=\color{blue!80!black}\bfseries,
  commentstyle=\color{green!50!black}\itshape,
  stringstyle=\color{red!70!black},
  numberstyle=\tiny\color{gray},
  numbers=left,
  numbersep=5pt,
  frame=single,
  framerule=0.4pt,
  rulecolor=\color{gray!40},
  backgroundcolor=\color{gray!4},
  breaklines=true,
  showstringspaces=false,
  tabsize=4,
  xleftmargin=1.5em,
  framexleftmargin=1.5em,
  aboveskip=0.6em,
  belowskip=0.4em,
  morekeywords={self,None,True,False,int,float,str},
}

\title{Duration Aware Scheduling for ASR Serving Under Workload Drift}

\author{Darshan Makwana\thanks{Correspondence to: Darshan Makwana (\href{mailto:darshan.makwana@sprinklr.com}{\color{blue}darshan.makwana@sprinklr.com})}\\
Sprinklr\\
Gurugram, India\\
\And
Yash Jogi\\
Sprinklr\\
Gurugram, India\\
\And
Harsh Kotta\\
Sprinklr\\
Gurugram, India\\
\And
Aayush Kubba\\
Sprinklr\\
Gurugram, India\\
}

\iclrfinalcopy 
\begin{document}

\maketitle

\begin{abstract}
Scheduling policies in large-scale Automatic Speech Recognition (ASR) serving pipelines play a key role in determining end-to-end (E2E) latency. Yet, widely used serving engines rely on first-come-first-served (FCFS) scheduling, which ignores variability in request duration and leads to head-of-line blocking under workload drift. We show that audio duration is an accurate proxy for job processing time in ASR models such as Whisper, and use this insight to enable duration-aware scheduling. We integrate two classical algorithms, Shortest Job First (SJF) and Highest Response Ratio Next (HRRN), into vLLM and evaluate them under realistic and drifted workloads. On LibriSpeech test-clean, compared to baseline, SJF reduces median E2E latency by up to $73\%$ at high load, but increases $90$th-percentile tail latency by up to $97\%$ due to starvation of long requests. HRRN addresses this trade-off: it reduces median E2E latency by up to $28\%$ while bounding tail-latency degradation to at most $24\%$. These gains persist under workload drift, with no throughput penalty and $<0.1$\,ms scheduling overhead per request.
\end{abstract}

\section{Introduction}

End-to-end latency is a critical quality-of-service metric for ASR systems. In interactive applications such as voice assistants \cite{li2017acoustic}, real-time captioning \cite{xue2022large}, and simultaneous interpretation \cite{ma2020simulmt}, users expect responses that are nearly instantaneous. Hence, any noticeable delay can reduce usability and overall user satisfaction.

\begin{figure}[ht]
\centering
\includegraphics[width=\textwidth]{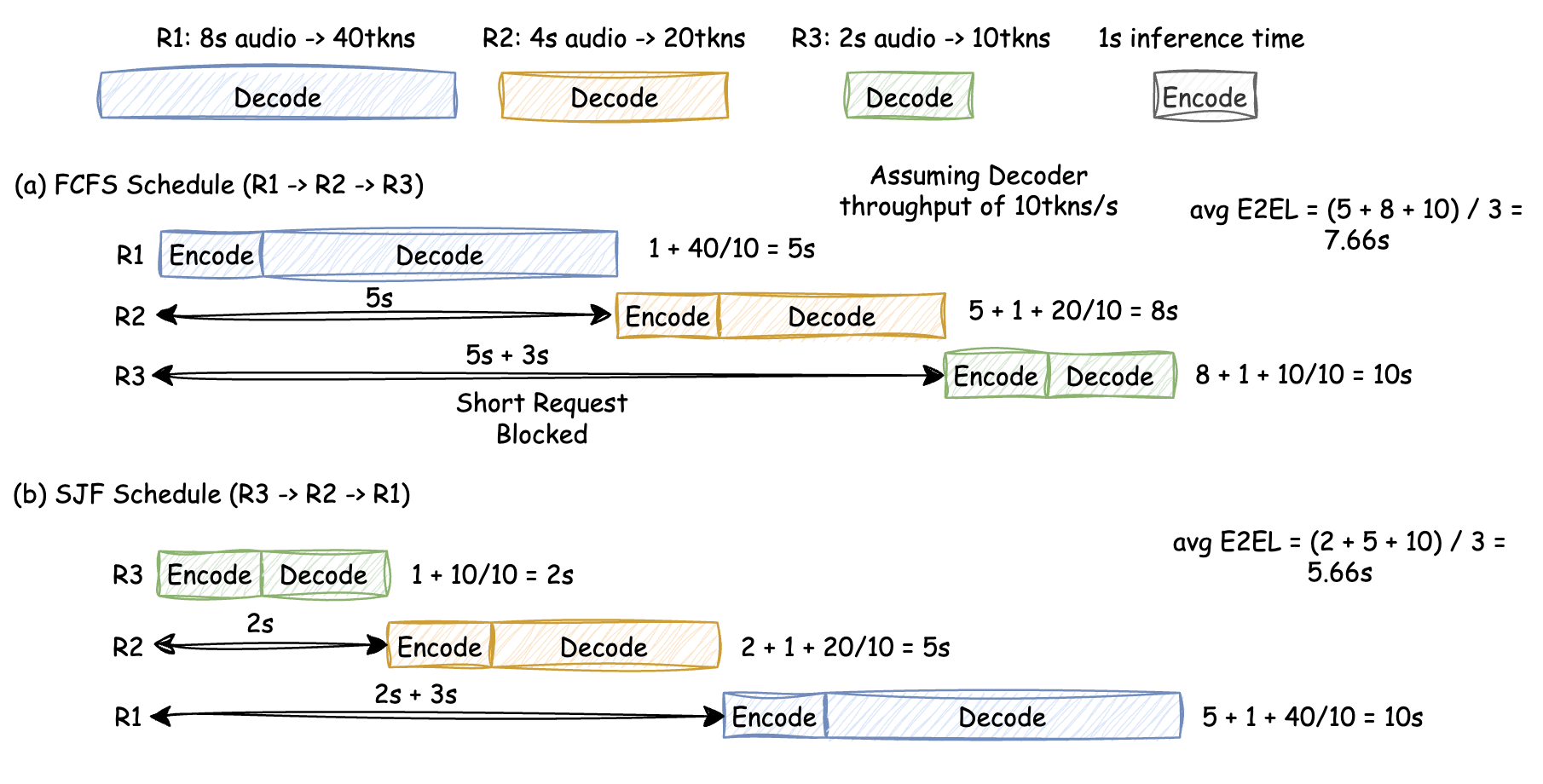}
\caption{Toy example illustrating head-of-line blocking under FCFS and the benefit of duration-aware scheduling. Three requests arrive in order $R_1,R_2,R_3$ with audio durations $8$~s, $4$~s, and $2$~s. We assume a constant encoder cost of $1$~s per request and a decoding rate of $5$ output tokens/s, yielding $40$, $20$, and $10$ output tokens, respectively. Under FCFS, serving $R_1$ first delays the two shorter requests and yields an average end-to-end latency of $(5{+}8{+}10)/3=7.66$~s. Reordering by shortest job first (SJF) reduces the average to $(2{+}5{+}10)/3=5.66$~s (a $1.4\times$ improvement). All numbers are illustrative and not intended to match measured system timings.}

\label{fig:scheduling_example}
\end{figure}

Popular speech-to-text (STT) models, such as Whisper \cite{radford2023robust}, are often deployed using inference engines like vLLM \cite{kwon2023efficient} and Orca \cite{yu2022orca}, which typically use a first-come-first-served (FCFS) scheduling policy. FCFS is simple and does not require any knowledge regarding the workload in advance. However, it can become inefficient under heavy load or when request distributions vary: long-running requests can block shorter ones, increasing queue delays and average latency. Figure~\ref{fig:scheduling_example} illustrates this with a simple example. In the given example, merely reordering requests by estimated job length reduces the average latency by around $1.4$ times.

To overcome the limitations of FCFS under variable workloads, scheduling algorithms that account for job processing time can produce more efficient execution orders and in turn reduce queuing delays. In this paper, we analyze and exploit the correlation between audio duration and job processing time to estimate job lengths and apply this estimation to two classical scheduling algorithms. The first is Shortest Job First (SJF) \cite{silberschatz2018operating}, which prioritizes shorter requests to minimize average waiting time. The second is Highest Response Ratio Next (HRRN) \cite{silberschatz2018operating}, which incorporates both waiting time and estimated job length to mitigate starvation of longer jobs.

We integrate the two algorithms, SJF and HRRN, into vLLM’s engine and evaluate them on two workloads: the LibriSpeech \cite{panayotov2015librispeech} \textit{test-clean} split and a synthetic workload derived from the LibriSpeech dataset with a uniform duration distribution, in order to study robustness under workload drift.

On the LibriSpeech dataset, we find that SJF, compared to baseline, reduces median end-to-end latency by up to $73\%$ and median time to first token by up to $93\%$ at high workload, with no throughput penalty. These gains persist across workloads: even on the synthetic split, SJF achieves up to a $67\%$ reduction in median end-to-end latency, confirming that the improvements arise from queue reordering rather than exploiting natural skew in audio durations present in the LibriSpeech dataset. Moreover, HRRN provides a practical alternative, delivering consistent median improvements while bounding $90th$-percentile tail latency degradation to at most $24\%$, compared to $97\%$ for SJF, compared to baseline. Additionally, both policies incur less than $0.1\,ms$ of scheduling overhead per request. Overall, our findings show that duration‑aware policies offer an effective, deployment‑ready enhancement for improving ASR responsiveness.

\section{Related Work}
\label{sec:related}

\paragraph{LLM serving systems.}
Modern inference engines such as Orca~\citep{yu2022orca} and vLLM~\citep{kwon2023efficient} introduced iteration-level scheduling and paged KV-cache management, forming the backbone of current serving stacks. Sarathi-Serve~\citep{agrawal2024taming} eliminates decode stalls via chunked prefill, while DistServe~\citep{zhong2024distserve} and Splitwise~\citep{patel2024splitwise} disaggregate prefill and decode phases across GPUs. All of these systems default to FCFS scheduling and are therefore susceptible to head-of-line blocking when request durations vary. Moreover, due to the architectural similarities between LLMs and Whisper, these serving frameworks are also commonly used for Whisper.

\paragraph{Scheduling for LLM inference.}
FastServe~\citep{wu2023fastserve} applies a skip-join multi-level feedback queue (MLFQ) to LLM serving, assigning requests to priority queues based on tokens generated so far. Fu~et~al.~\citep{fu2024efficient} train a small ranking model to predict relative output lengths and approximate SJF ordering, achieving up to $2.8\times$ lower latency on chatbot workloads. Andes~\citep{liu2024andes} introduces a quality-of-experience metric and schedules at token granularity to maximize user satisfaction. Sheng~et~al.~\citep{sheng2024fairness} propose a Virtual Token Counter for fair scheduling that accounts for both input and output tokens. These methods all target text-based LLM workloads where output length is unknown a priori and must be predicted or estimated online.

\paragraph{Output length prediction.}
Several works predict LLM output lengths to improve scheduling or memory allocation. S3~\citep{jin2023s3} uses a classification model to predict sequence lengths for KV-cache sizing. Zheng~et~al.~\citep{zheng2023response} let the LLM itself predict its response length before generation. Magnus~\citep{cheng2024magnus} combines semantic features with HRRN scheduling. Qiu~et~al.~\citep{qiu2024efficient} train a BERT proxy model for speculative SJF. All of these approaches incur non-trivial overhead: either an auxiliary predictor that consumes GPU cycles, or self-prediction tokens that delay the actual response. We draw inspiration from these works and adapt these ideas to the problem of ASR output length prediction.

\paragraph{Size-based scheduling.}
Prioritizing shorter jobs originates in classical scheduling theory~\citep{silberschatz2018operating}. Harchol-Balter~et~al.~\citep{harchol2003size} show that Shortest Remaining Processing Time (SRPT) scheduling yields large mean response time reductions in web servers when file sizes are known, with limited unfairness under heavy-tailed workloads~\citep{bansal2001analysis}. Clockwork~\citep{gujarati2020clockwork} exploits deterministic DNN execution times for predictable latency. Our work draws on this tradition: in ASR serving, audio duration plays the role of file size a freely available signal for job length that requires no prediction.

The central difference between our approach and prior scheduling work for LLM inference is the source of the job-length signal. For text-based LLMs, output length is fundamentally unpredictable, forcing systems to train auxiliary models~\citep{fu2024efficient,jin2023s3,qiu2024efficient}, prompt the LLM itself~\citep{zheng2023response}, or rely on heuristic features~\citep{cheng2024magnus}. In ASR workloads, audio duration is known at request arrival and correlates strongly with processing time (\S\ref{sec:methodology}), providing a practically zero-overhead scheduling signal that is trivially deployable solution.

\section{Methodology}
\label{sec:methodology}

This section first analyzes the relationship between audio duration and job processing time, followed by description of the two scheduling algorithms considered in our study.

\subsection{Correlation between audio duration and job processing time}

\begin{figure}[t]
\centering
\includegraphics[width=\textwidth]{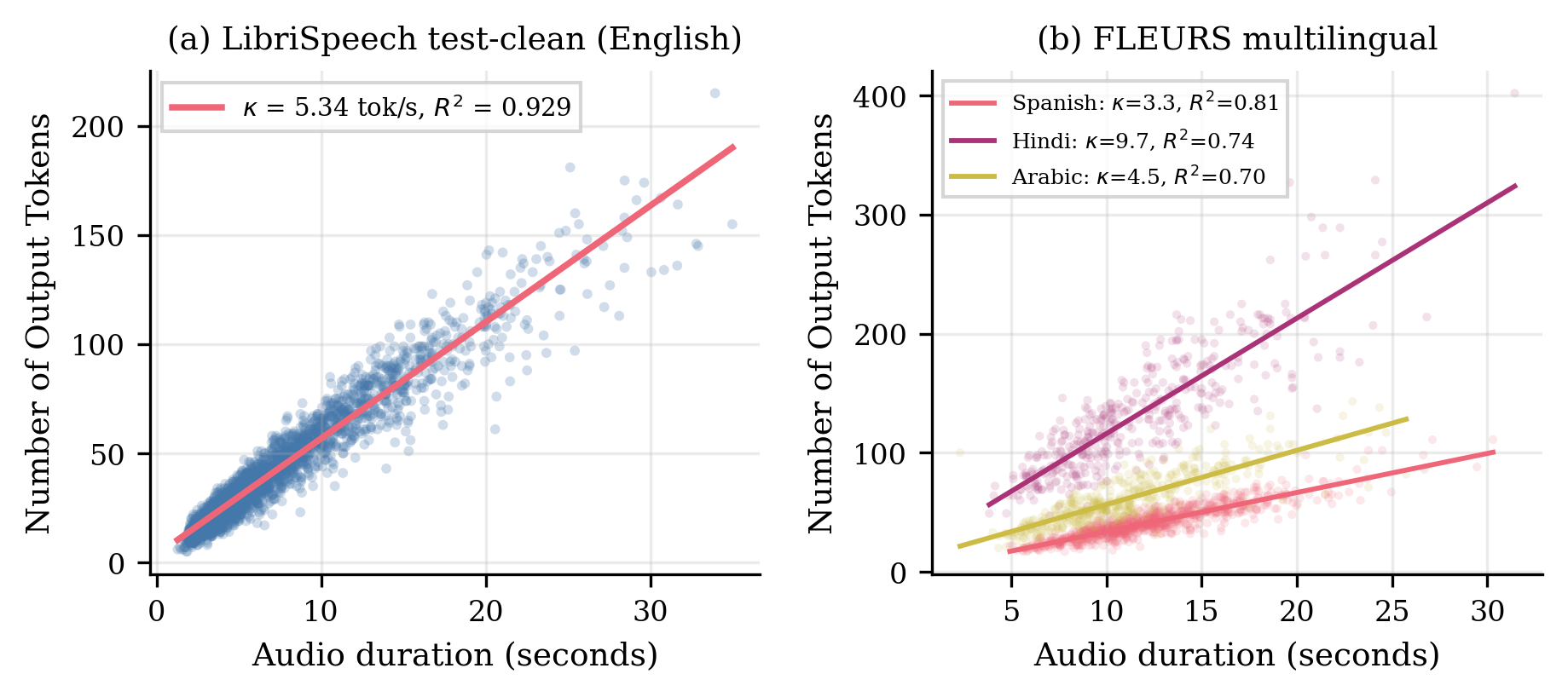}
\caption{Scatter plots showing the relationship between audio duration and ASR output token count. (a) On the LibriSpeech English test set, token count increases linearly with audio duration, indicating a strong correlation. (b) On the FLEURS test sets for Spanish, Hindi, and Arabic, the linear duration–token relationship is maintained, demonstrating that this correlation generalizes across typologically diverse languages.}

\label{fig:correlation}
\end{figure}

In encoder–decoder STT models such as Whisper, an input audio is processed by the Whisper encoder in fixed-length segments of $30$ seconds, leading to near-constant encoding time per segment. Decoding time, however, grows approximately linearly with the number of generated output tokens. As a result, overall processing time is dominated by decoding and is largely determined by the number of generated output tokens.

In ASR models, the number of generated output tokens is strongly correlated with the duration of the input audio. This relationship arises from the relatively stable rate of human speech, whereby longer audio segments contain more words and therefore more tokens. We empirically analyze this correlation using the LibriSpeech test-clean set for English and the FLEURS dataset for Spanish, Hindi, and Arabic. Transcriptions are generated using the \texttt{whisper-large-v3} ASR model. Figure~\ref{fig:correlation} presents scatter plots of audio duration versus output token count across all the datasets, revealing a clear linear relationship, which can be expressed as:

\begin{equation}
\hat{n} = d \times \kappa
\label{eq:estimation}
\end{equation}

where $d$ is audio duration, $\kappa$ is a language-specific constant, and $\hat{n}$ is the estimated number of output tokens. The values of $\kappa$ for English, Arabic, and Spanish are reported in Figure~\ref{fig:correlation} along with their estimation errors. Since job processing time is proportional to token count, audio duration provides a reliable estimate of processing time. Because scheduling algorithms depend only on the relative ordering of jobs, exact job time estimates are not required. Therefore, we use audio duration as a proxy for job processing time in the scheduling algorithms described below.

\subsection{Scheduling Policies}

\paragraph{Shortest Job First (SJF).} 
SJF prioritizes shorter jobs from the pool of incoming requests, which minimizes the average waiting time across all jobs~\citep{silberschatz2018operating}. If multiple jobs have the same duration, earlier arrivals are prioritized. We implement SJF using a min-heap, which has \(O(\log n)\) cost for both insertion and deletion. 

Although SJF does reduce average waiting time, it can cause starvation for long jobs when short jobs continuously arrive, as long requests may be delayed indefinitely~\citep{silberschatz2018operating,bansal2001analysis}. 

\paragraph{Highest Response Ratio Next (HRRN).} 

To address the starvation issue inherent in SJF, HRRN incorporates both waiting time and estimated job duration when making scheduling decisions. Jobs are ordered according to their response ratio, defined as:
\[
\text{Response Ratio} = \frac{\text{Waiting Time} + \text{Estimated Job Time}}{\text{Estimated Job Time}}.
\]
The estimated job time in the above equation is derived from the audio duration using the linear relationship in Equation~\ref{eq:estimation}. Because HRRN only requires relative ordering, the exact token count need not be precise, audio duration $d$ can be used directly as the estimated job time. Jobs with higher response ratios are scheduled first, allowing requests that have waited longer to gradually gain priority while still favoring shorter jobs~\citep{silberschatz2018operating}.

\section{Experiments}
\label{sec:experiments}

\subsection{Setup}

For all experiments, we use the Whisper model as the ASR model, specifically the \texttt{whisper-large-v3} variant with $1.5B$ parameters. The model is served using vLLM with the \texttt{openai-audio} backend, with maximum output length of $448$ tokens, and chunked prefill~\citep{agrawal2024taming} enabled. We configure vLLM with a maximum batch size of $256$ (the default setting) and a GPU memory utilization target of $95\%$. All experiments are conducted on a single NVIDIA $A100$ GPU with $40$ GB HBM2e memory, running on CUDA $12.1$. To study how scheduling policies perform under workloads with different characteristics, we use two datasets derived from the LibriSpeech test-clean subset:

\paragraph{Original LibriSpeech split.} This is Librispeech as it is with audio durations ranging from $1$ to $35$\,s with $\mu{=}7.4$\,s and $\sigma{=}5.15$\,s. The distribution is right-skewed, with the majority of utterances being short, reflecting realistic ASR serving conditions. Since the dataset has a right-skewed duration distribution, where the majority of utterances are short, precisely the regime where SJF thrives, because the queue is dominated by quick jobs that SJF can rapidly drain. A natural concern is whether the observed gains are an artifact of this skew: perhaps SJF only helps when most requests are already short and a few long outliers cause head-of-line blocking.

\paragraph{Synthetic LibriSpeech Split. } To disentangle the effect of duration \emph{ordering} from the effect of duration \emph{distribution shape}, we construct a \emph{synthetic split} from six LibriSpeech with the exact target durations $\{5, 10, 15, 20, 25, 30\}$\,s. This produces a discrete-uniform distribution with higher mean ($\mu{=}17.5$\,s) and higher variance ($\sigma{\approx}8.5$\,s) than LibriSpeech. Crucially, this removes the right-skew: short and long jobs are now equally likely, and the mean job is substantially longer. If scheduling gains collapse under this flat, heavy workload, we would conclude the benefit was distribution-dependent. If the gains persist, this indicates that priority reordering is valuable and that the improvements observed on LibriSpeech are not solely attributable to its right-skewed duration distribution.

We simulate a realistic serving scenario in which audio requests arrive as a continuous stream following a Poisson process with a burstiness factor of $1.0$. To evaluate performance under varying workload intensities, we vary the request arrival rate from $1$ to $25$ requests per second. For each configuration, the simulation is run for $5$ minutes. To ensure a fair comparison, all scheduling policies are evaluated using identical arrival sequences.

\paragraph{Evaluation Metrics.} We evaluate the policies using metrics that reflect system performance and user experience. End-to-end latency (E2EL) measures the total time from request arrival to transcription completion, including queuing, prefill, and decode phases. Time to first token (TTFT) captures the delay until the first output appears, reflecting perceived responsiveness. We report the $50th$ percentile ($P50$) and $90th$ percentile ($P90$) for each metric. $P50$ captures the typical user experience, which is the latency at or below which half of all requests complete, while $P90$ reveals tail behavior, quantifying how the slowest requests are affected by the scheduling policy. Reporting both is critical for evaluating scheduling trade-offs: a policy that dramatically improves $P50$ may do so at the expense of $P90$, trading better median experience for degraded worst-case reliability.

\subsection{Results}

\begin{table}[t]
\caption{LibriSpeech test-clean: Latency comparison across request rates (ms). Parenthesized values show relative change versus FCFS. Bold values indicate the highest value among the three policies.\colorbox{green!12}{Green}: $>$3\% improvement; \colorbox{red!12}{red}: $>$3\% degradation.}
\label{tab:librispeech}
\vspace{0.5em}
\centering
\footnotesize
\setlength{\tabcolsep}{2.5pt}
\resizebox{\textwidth}{!}{%
\begin{tabular}{c ccc ccc ccc}
\toprule
\multirow{2}{*}{\textbf{Rate}} & \multicolumn{3}{c}{\textbf{P50 TTFT (ms) $\downarrow$}} & \multicolumn{3}{c}{\textbf{P50 E2EL (ms) $\downarrow$}} & \multicolumn{3}{c}{\textbf{P90 E2EL (ms) $\downarrow$}} \\
\cmidrule(lr){2-4} \cmidrule(lr){5-7} \cmidrule(lr){8-10}
& FCFS & SJF & HRRN & FCFS & SJF & HRRN & FCFS & SJF & HRRN \\
\midrule
1 & 64.6 & \cellcolor{green!12}\textbf{61.5} {\scriptsize($-$5\%)} & \cellcolor{red!12}73.3 {\scriptsize(+13\%)} & 88.2 & \cellcolor{green!12}\textbf{84.9} {\scriptsize($-$4\%)} & \cellcolor{red!12}96.4 {\scriptsize(+9\%)} & 143.8 & \textbf{141.6} {\scriptsize($-$2\%)} & \cellcolor{red!12}155.7 {\scriptsize(+8\%)} \\
5 & 64.7 & \textbf{62.8} {\scriptsize($-$3\%)} & 64.4 & 112.4 & \cellcolor{green!12}\textbf{108.9} {\scriptsize($-$3\%)} & 112.5 & 276.7 & \cellcolor{green!12}\textbf{261.3} {\scriptsize($-$6\%)} & 275.9 \\
10 & 118.6 & 115.5 {\scriptsize($-$3\%)} & \textbf{115.1} {\scriptsize($-$3\%)} & 213.0 & \cellcolor{green!12}200.8 {\scriptsize($-$6\%)} & \cellcolor{green!12}\textbf{195.6} {\scriptsize($-$8\%)} & 742.7 & \cellcolor{green!12}682.5 {\scriptsize($-$8\%)} & \cellcolor{green!12}\textbf{681.7} {\scriptsize($-$8\%)} \\
\midrule
15 & 192.0 & \cellcolor{green!12}\textbf{144.2} {\scriptsize($-$25\%)} & \cellcolor{red!12}198.3 {\scriptsize(+3\%)} & 547.9 & \cellcolor{green!12}\textbf{518.3} {\scriptsize($-$5\%)} & \cellcolor{red!12}598.3 {\scriptsize(+9\%)} & 1561.6 & \cellcolor{green!12}\textbf{1467.9} {\scriptsize($-$6\%)} & \cellcolor{red!12}1615.4 {\scriptsize(+3\%)} \\
20 & 1467.8 & \cellcolor{green!12}\textbf{193.3} {\scriptsize($-$87\%)} & \cellcolor{green!12}1173.9 {\scriptsize($-$20\%)} & 2728.7 & \cellcolor{green!12}\textbf{1325.8} {\scriptsize($-$51\%)} & \cellcolor{green!12}2231.2 {\scriptsize($-$18\%)} & \textbf{3954.8} & \cellcolor{red!12}7094.6 {\scriptsize(+79\%)} & \cellcolor{red!12}4996.6 {\scriptsize(+26\%)} \\
25 & 4273.4 & \cellcolor{green!12}\textbf{296.3} {\scriptsize($-$93\%)} & \cellcolor{green!12}2867.3 {\scriptsize($-$33\%)} & 5423.8 & \cellcolor{green!12}\textbf{1486.8} {\scriptsize($-$73\%)} & \cellcolor{green!12}3897.9 {\scriptsize($-$28\%)} & \textbf{8334.4} & \cellcolor{red!12}16405.3 {\scriptsize(+97\%)} & \cellcolor{red!12}10347.3 {\scriptsize(+24\%)} \\
\bottomrule
\end{tabular}}
\end{table}

\begin{figure}[t]
\centering
\includegraphics[width=\textwidth]{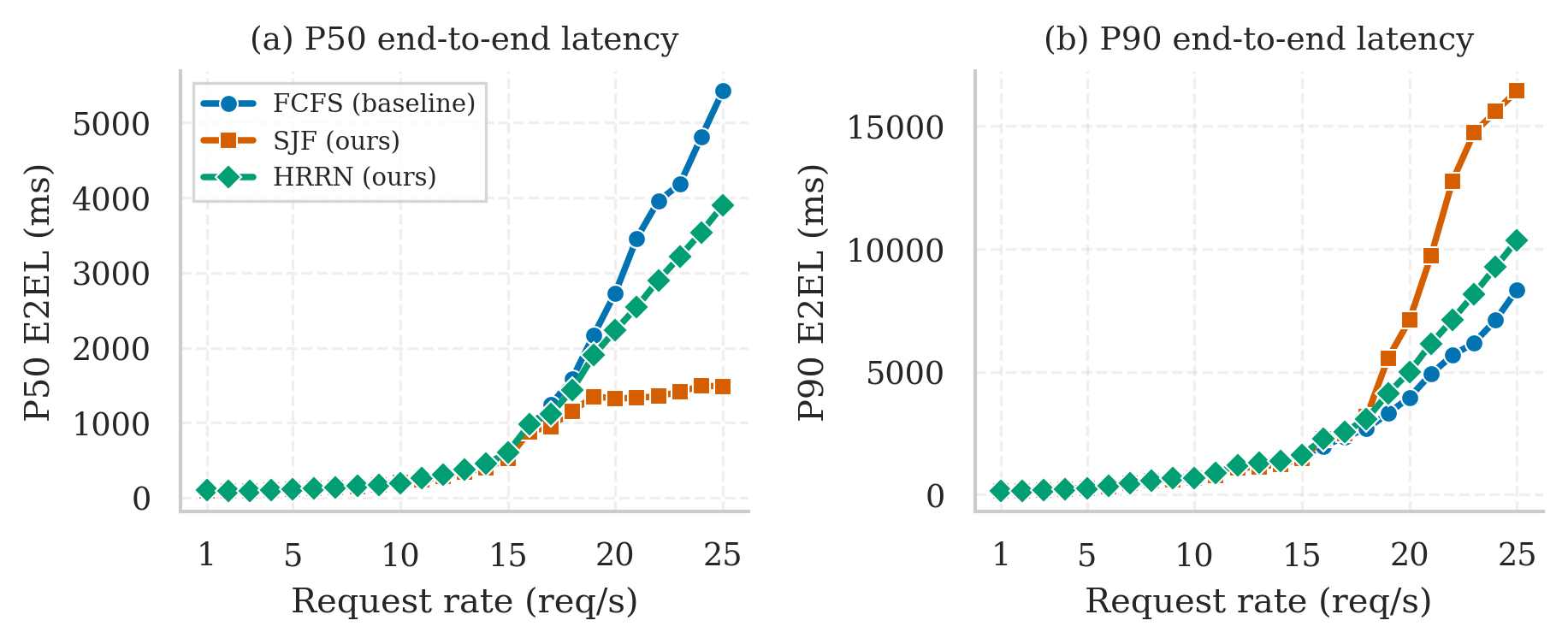}
\caption{LibriSpeech test-clean: End-to-end latency scaling ($1$--$25$ req/s). (a) SJF reduces $P50$ E2EL by up to $73\%$ at $25$~req/s, (b) while $P90$ E2EL reveals the starvation trade-off at high load.}
\label{fig:ls_latency}
\end{figure}

\begin{figure}[t]
\centering
\includegraphics[width=\textwidth]{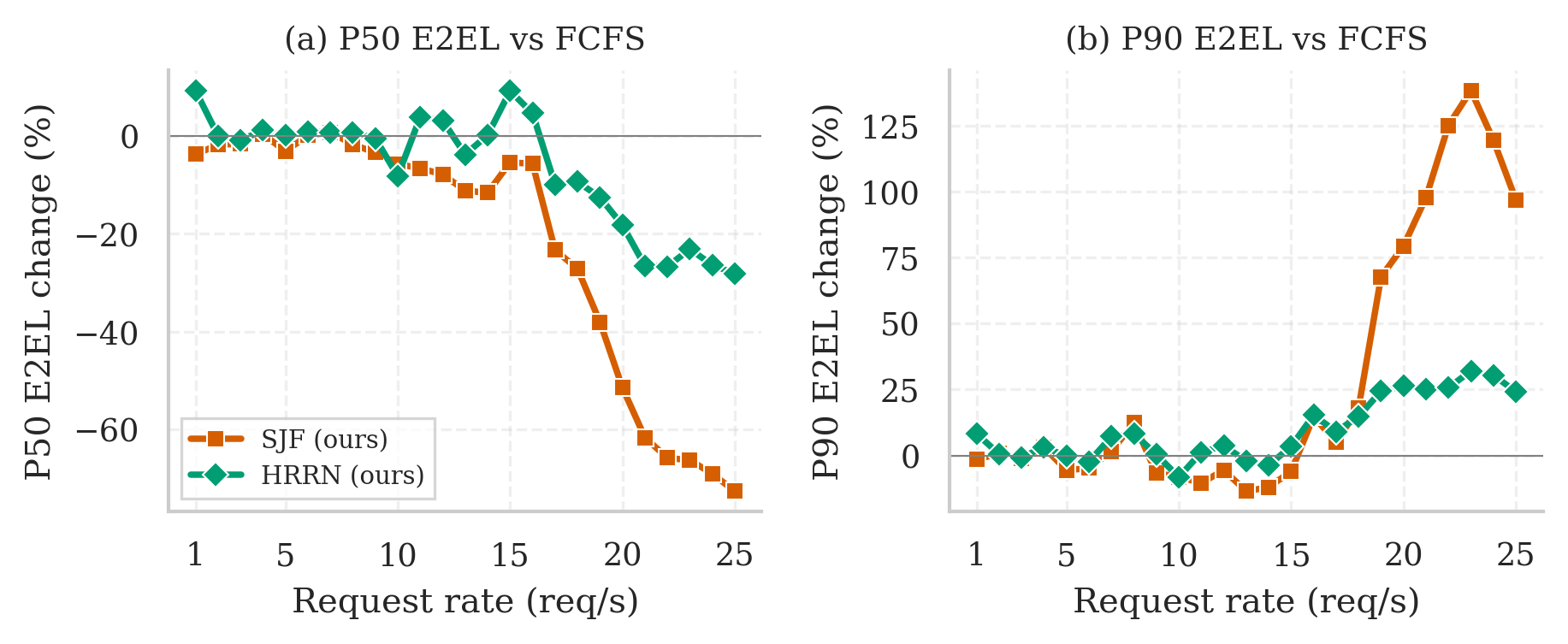}
\caption{LibriSpeech test-clean: Percentage change in E2EL versus FCFS ($1$--$25$~req/s). Negative values indicate improvement. SJF's $P50$ gains deepen monotonically with load, while $P90$ degradation emerges beyond $15$~req/s.}
\label{fig:ls_relative}
\end{figure}

Figure~\ref{fig:ls_latency} and~\ref{fig:ls_relative} show the performance of the three scheduling policies on the original LibriSpeech split. To examine the performance under different workload conditions, we divide the workload into three request‑rate ranges: low ($1$–$10$ req/s), moderate ($10$–$20$ req/s), and heavy ($20$–$25$ req/s). The paragraphs below summarize each policy’s performance in these ranges; a table of the numerical results appears in Table~\ref{tab:librispeech}. All percentage comparisons done below are made relative to FCFS.

Under low load, resources are sufficient to handle incoming requests and queues rarely form, so all policies perform similarly. However, at the upper end of this range at $10$~req/s, both SJF and HRRN show measurable improvements: SJF reduces $P50$ and $P90$ E2EL by $6\%$ and $8\%$, respectively, while HRRN achieves reductions of $8\%$ for both metrics. This indicates that even when the system is lightly loaded, reordering requests according to estimated job size can yield latency benefits.

As the request rate enters the moderate range and approaches the system’s service capacity, queues begin to lengthen, making SJF’s advantages more pronounced. At $17$~req/s, SJF reduces $P50$ E2EL by $23$\%, though $P90$ E2EL begins to show the starvation trade-off with a modest $5\%$ increase. HRRN lowers $P50$ E2EL by $10\%$ while its $P90$ E2EL increases by $9\%$. HRRN's weaker P50 improvement relative to SJF reflects its aging mechanism: by gradually promoting long‑waiting jobs, it sacrifices some median-latency gain to prevent starvation a trade-off that becomes more consequential as queue depth grows. The moderate-load regime thus represents a practical sweet spot for SJF: the system is sufficiently busy to exploit job reordering, yet not saturated enough to cause severe starvation for longer jobs.

Under heavy load, SJF's improvements become even more pronounced. At $23$~req/s, SJF reduces $P50$ E2EL by $66\%$. However, this comes at a significant cost for tail latency, as $P90$ E2EL increases by $138\%$. In this case, HRRN provides a practical middle ground: it reduces $P50$ E2EL only by $23\%$ while limiting $P90$ E2EL degradation to $32\%$.Moreover, results for TTFT, a metric more directly reflective of queueing delay, which show even greater improvements, are provided in Appendix~\ref{app:ttft}.

\begin{figure}[t]
\centering
\includegraphics[width=\textwidth]{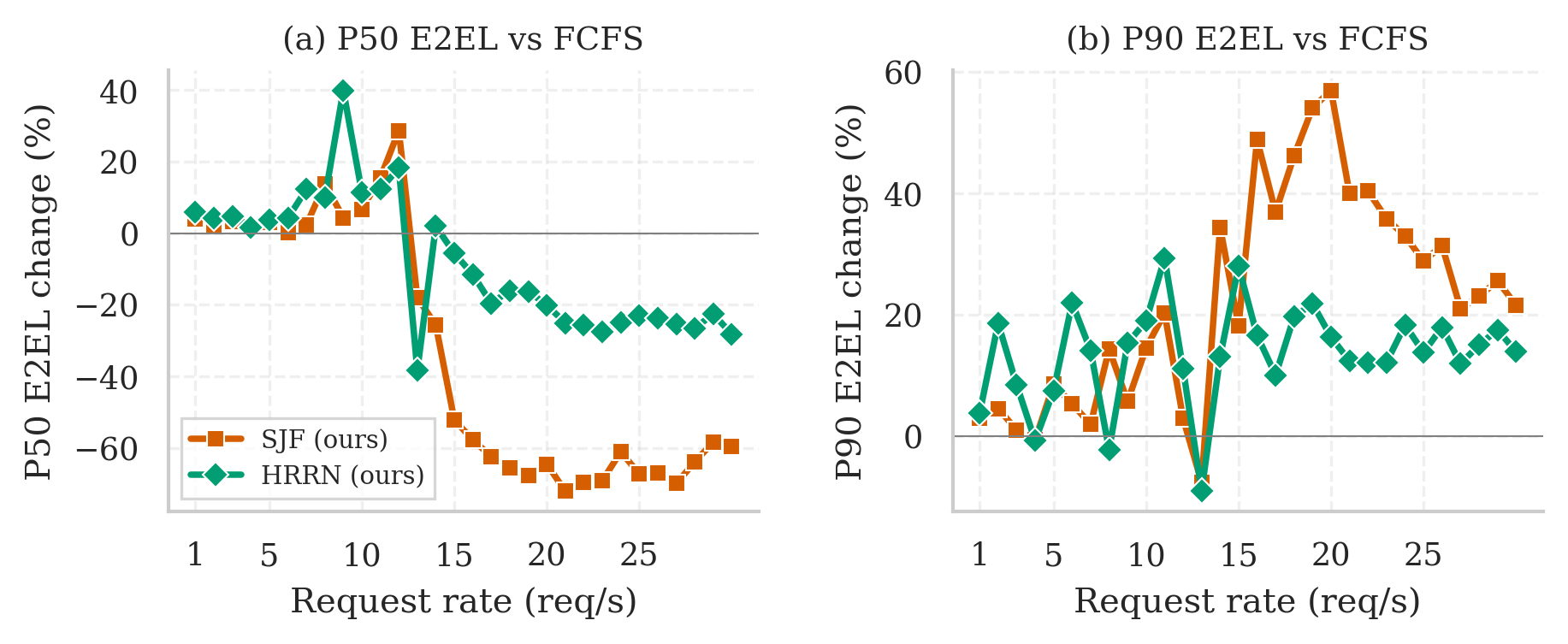}
\caption{Synthetic split: Percentage change in E2EL versus FCFS ($1$--$30$~req/s). SJF's E2EL gains match LibriSpeech, confirming reordering benefits are not an artifact of the natural duration skew.}
\label{fig:ctrl_relative}
\end{figure}

Figure~\ref{fig:ctrl_latency} and~\ref{fig:ctrl_relative} present the same evaluation on the synthetic split. Table~\ref{tab:synthetic} reports the corresponding numerical results. The two main findings are:

\begin{table}[t]
\caption{Synthetic split: Latency comparison across request rates (ms). Parenthesized values show relative change versus FCFS. Bold values indicate the highest value among the three policies. \colorbox{green!12}{Green}: $>3\%$ improvement; \colorbox{red!12}{red}: $>3\%$ degradation.}
\label{tab:synthetic}
\vspace{0.5em}
\centering
\footnotesize
\setlength{\tabcolsep}{2.5pt}
\resizebox{\textwidth}{!}{%
\begin{tabular}{c ccc ccc ccc}
\toprule
\multirow{2}{*}{\textbf{Rate}} & \multicolumn{3}{c}{\textbf{P50 TTFT (ms) $\downarrow$}} & \multicolumn{3}{c}{\textbf{P50 E2EL (ms) $\downarrow$}} & \multicolumn{3}{c}{\textbf{P90 E2EL (ms) $\downarrow$}} \\
\cmidrule(lr){2-4} \cmidrule(lr){5-7} \cmidrule(lr){8-10}
& FCFS & SJF & HRRN & FCFS & SJF & HRRN & FCFS & SJF & HRRN \\
\midrule
1 & 64.0 & \cellcolor{red!12}67.1 {\scriptsize(+5\%)} & \cellcolor{red!12}68.6 {\scriptsize(+7\%)} & 84.8 & \cellcolor{red!12}88.3 {\scriptsize(+4\%)} & \cellcolor{red!12}89.9 {\scriptsize(+6\%)} & 158.0 & 162.8 {\scriptsize(+3\%)} & \cellcolor{red!12}164.0 {\scriptsize(+4\%)} \\
5 & 66.0 & \cellcolor{red!12}68.4 {\scriptsize(+4\%)} & \cellcolor{red!12}68.8 {\scriptsize(+4\%)} & 117.7 & 121.1 {\scriptsize(+3\%)} & \cellcolor{red!12}122.1 {\scriptsize(+4\%)} & 242.2 & \cellcolor{red!12}262.9 {\scriptsize(+9\%)} & \cellcolor{red!12}260.3 {\scriptsize(+7\%)} \\
10 & 120.3 & \cellcolor{red!12}126.8 {\scriptsize(+5\%)} & 124.3 {\scriptsize(+3\%)} & 214.8 & \cellcolor{red!12}229.3 {\scriptsize(+7\%)} & \cellcolor{red!12}239.5 {\scriptsize(+11\%)} & 741.1 & \cellcolor{red!12}849.0 {\scriptsize(+15\%)} & \cellcolor{red!12}881.6 {\scriptsize(+19\%)} \\
\midrule
15 & 2654.0 & \cellcolor{green!12}\textbf{267.1} {\scriptsize($-$90\%)} & \cellcolor{green!12}2263.5 {\scriptsize($-$15\%)} & 4788.6 & \cellcolor{green!12}\textbf{2287.3} {\scriptsize($-$52\%)} & \cellcolor{green!12}4520.1 {\scriptsize($-$6\%)} & \textbf{7529.9} & \cellcolor{red!12}8902.4 {\scriptsize(+18\%)} & \cellcolor{red!12}9635.5 {\scriptsize(+28\%)} \\
20 & 6661.2 & \cellcolor{green!12}\textbf{906.7} {\scriptsize($-$86\%)} & \cellcolor{green!12}4702.7 {\scriptsize($-$29\%)} & 8686.9 & \cellcolor{green!12}\textbf{3079.7} {\scriptsize($-$65\%)} & \cellcolor{green!12}6939.6 {\scriptsize($-$20\%)} & \textbf{14325.2} & \cellcolor{red!12}22478.8 {\scriptsize(+57\%)} & \cellcolor{red!12}16664.3 {\scriptsize(+16\%)} \\
25 & 9923.9 & \cellcolor{green!12}\textbf{1574.4} {\scriptsize($-$84\%)} & \cellcolor{green!12}6577.1 {\scriptsize($-$34\%)} & 12123.3 & \cellcolor{green!12}\textbf{3966.3} {\scriptsize($-$67\%)} & \cellcolor{green!12}9340.9 {\scriptsize($-$23\%)} & \textbf{19664.3} & \cellcolor{red!12}25336.7 {\scriptsize(+29\%)} & \cellcolor{red!12}22378.3 {\scriptsize(+14\%)} \\
\bottomrule
\end{tabular}}
\end{table}

\begin{figure}[t]
\centering
\includegraphics[width=\textwidth]{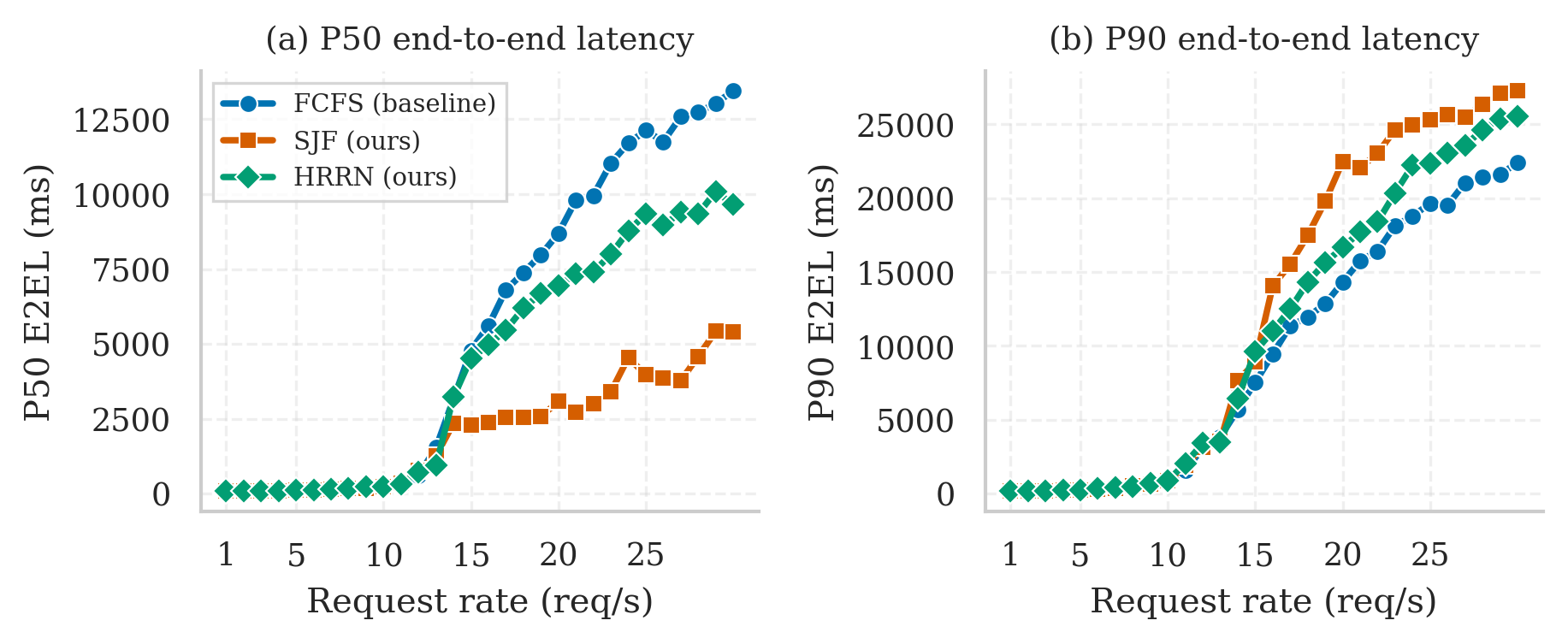}
\caption{Synthetic split: End-to-end latency scaling ($1$--$30$ req/s). (a) SJF's $P50$ E2EL advantage ($-67\%$ at $25$~req/s) persists under a uniform duration distribution, (b) while $P90$ E2EL degradation is moderated compared to LibriSpeech's right-skewed workload.}
\label{fig:ctrl_latency}
\end{figure}

\textbf{SJF E2EL gains persist.} At $25$~req/s, SJF reduces $P50$ E2EL by $67\%$. At $20$~req/s, SJF delivers $65\%$ P50 E2EL reduction, showing that end-to-end latency benefits generalizes across this workload distribution as well, which has a more uniform duration distribution than the LibriSpeech test-clean dataset.

\textbf{Tail penalty is moderated.} Despite the wider duration range, SJF's tail cost is lower than on LibriSpeech: at $25$~req/s, $P90$ E2EL inflation is $29\%$ versus $97\%$; at $20$~req/s, $57\%$ versus $79\%$. HRRN's tail penalty also shrinks ($14\%$ versus $24\%$ at $25$~req/s). The explanation is distributional: LibriSpeech's right-skew means the queue is flooded with short requests that SJF continuously prioritizes, starving the few long outliers indefinitely. Under a uniform distribution, long requests arrive at the same frequency as short ones, so the queue is less persistently dominated by short jobs and starvation episodes are shorter-lived.

\begin{figure}[t]
\centering
\includegraphics[width=\textwidth]{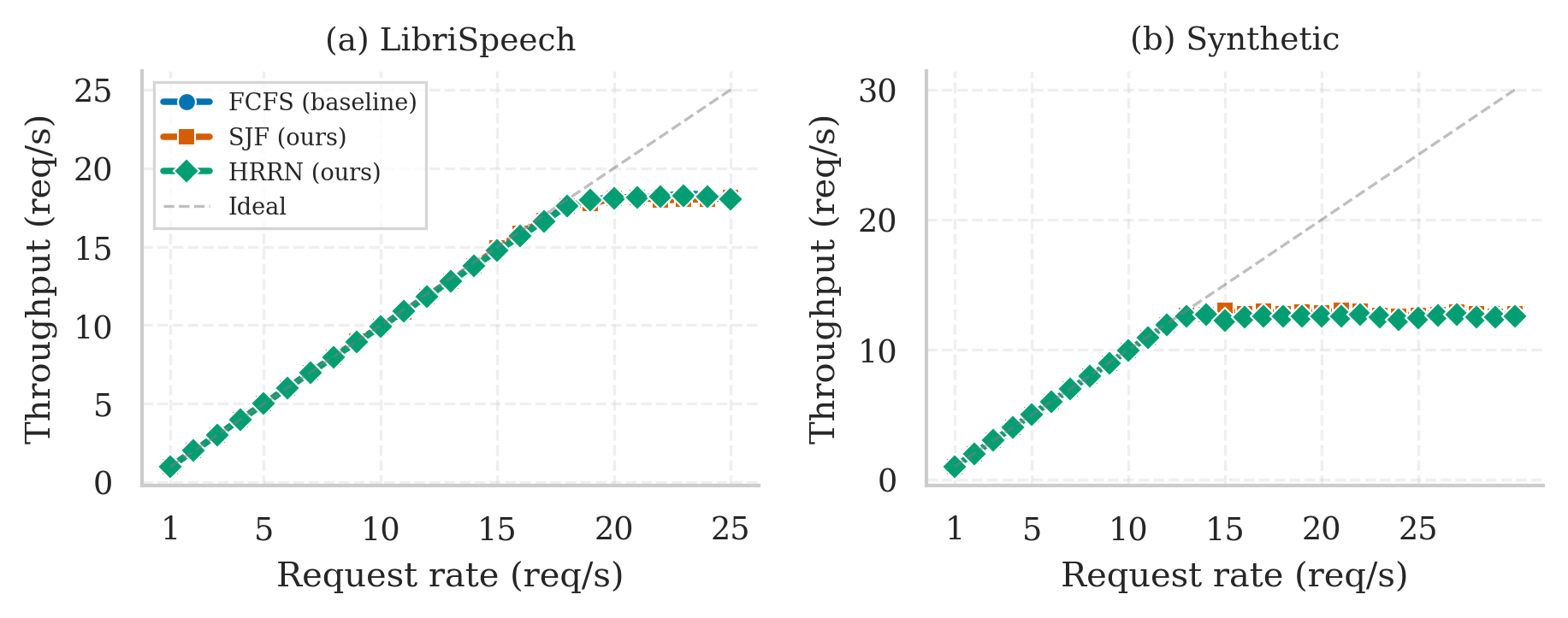}
\caption{Request throughput for both workloads. All scheduling policies achieve identical throughput, saturating at $\sim 18$ req/s on LibriSpeech and $\sim 13$ req/s on the synthetic split, confirming negligible scheduling overhead.}
\label{fig:throughput}
\end{figure}

Another important consideration is whether scheduling improvements affect overall throughput of the system. Figure~\ref{fig:throughput} presents the request rate versus throughput for all three policies. All of them achieve identical request throughput on both datasets. There is no measurable difference because the only overhead is the scheduling decision, which adds on average $<0.1$\,ms per request, which is negligible compared to the ${\sim}60-100$\,ms processing time for one decoder step in ASR.

\section{Limitations and Future Work}

\paragraph{Silence sensitivity.}
Our duration-based estimator assumes that spoken content fills most of the audio clip. However, when recordings contain extended silence segments (e.g., pauses, leading/trailing silence), the estimator can overestimate the number of output tokens, since silent regions produce no transcription tokens. A natural mitigation is to apply a voice activity detection (VAD) module as a preprocessing step, computing $d$ as the sum of speech-active segments rather than the raw file duration. Lightweight VAD models (e.g., Silero VAD~\citep{silero2024vad}) add $<5ms$ per utterance and could be integrated at the API endpoint with negligible overhead.

\paragraph{Adaptive $\kappa$.}
The current system employs a fixed $\kappa$ across all requests. In multilingual deployments, where languages exhibit different tokenization densities, as shown in Figure~\ref{fig:correlation}, this assumption may be limiting and can lead to suboptimal estimation accuracy.

\paragraph{Dynamic policy switching.}
Our experiments evaluate each scheduling policy in isolation. A production system could monitor queue depth, tail-latency percentiles, and starvation indicators in real time, dynamically switching between FCFS, SJF, and HRRN to match the current load regime.

\section{Conclusion}
\vspace{-0.5ex}
We show that first-come-first-served scheduling has clear limitations for ASR serving when request durations vary, and that audio duration is a reliable, zero-overhead proxy for job length in encoder--decoder models such as Whisper. After integrating SJF and HRRN into vLLM, we observe up to $73\%$ lower median E2E latency and up to $93\%$ lower median TTFT on LibriSpeech at high load. These improvements persist under workload drift, while HRRN bounds tail-latency degradation to at most $24\%$. Duration-aware scheduling is a simple, deployment-ready lever for improving user-perceived latency in production ASR systems.

\bibliography{iclr2026_conference}
\bibliographystyle{iclr2026_conference}

\appendix
\section{Appendix}
\subsection{Time to First Token Analysis}
\label{app:ttft}

While the main text focuses on end-to-end latency (E2EL), time to first token (TTFT) provides complementary insight into queuing behavior. Because TTFT measures only the delay before decoding begins, it isolates the scheduling decision's direct impact on queue wait time.

\begin{figure}[h]
\centering
\includegraphics[width=\textwidth]{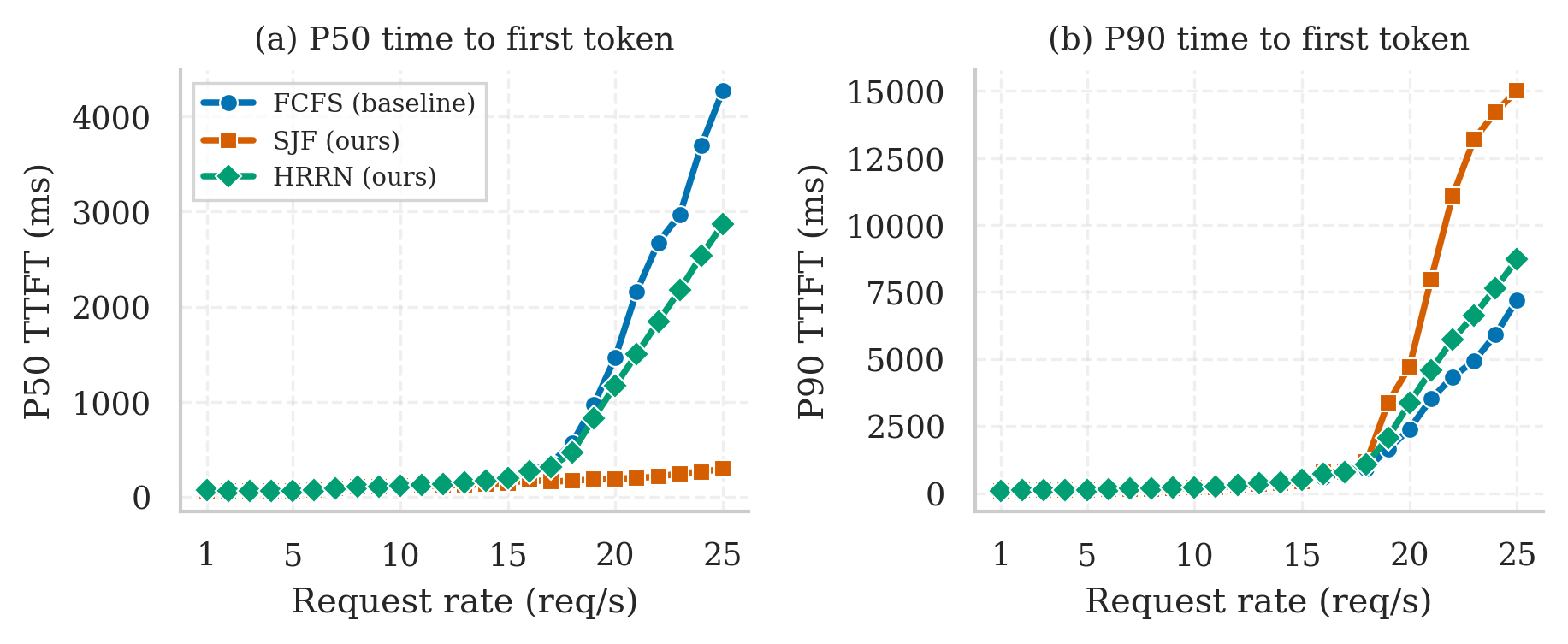}
\caption{LibriSpeech TTFT scaling (1--25 req/s). SJF keeps $P50$ TTFT below $300$\,ms at all rates while FCFS exceeds $4$\,s.}
\label{fig:ls_ttft}
\end{figure}

\begin{figure}[h]
\centering
\includegraphics[width=\textwidth]{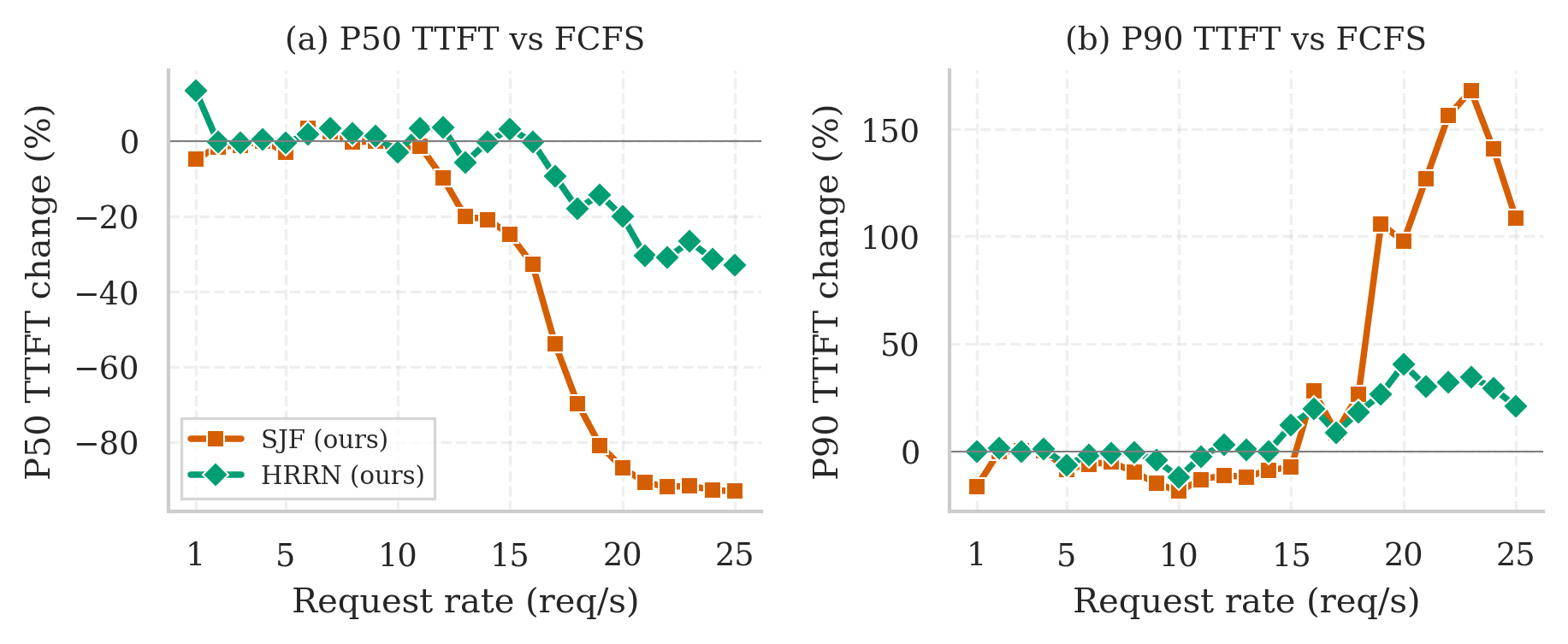}
\caption{LibriSpeech: percentage change in TTFT versus FCFS ($1$--$25$~req/s). SJF's $P50$ TTFT gains deepen monotonically, reaching $-93\%$ at $25$~req/s.}
\label{fig:ls_ttft_rel}
\end{figure}

Figure~\ref{fig:ls_ttft} shows TTFT scaling on LibriSpeech. At heavy load (20--25~req/s), SJF reduces $P50$ TTFT by up to $93\%$ from $4273$\,ms under FCFS to $296$\,ms, demonstrating that priority scheduling effectively reduces queueing delay for the majority of requests. Even at the $P90$ percentile, SJF maintains sub-second TTFT up to $15$~req/s. HRRN achieves $-33\%$ $P50$ TTFT at $25$~req/s while limiting $P90$ TTFT degradation relative to SJF. The corresponding percentage changes are shown in Figure~\ref{fig:ls_ttft_rel}.

\begin{figure}[h]
\centering
\includegraphics[width=\textwidth]{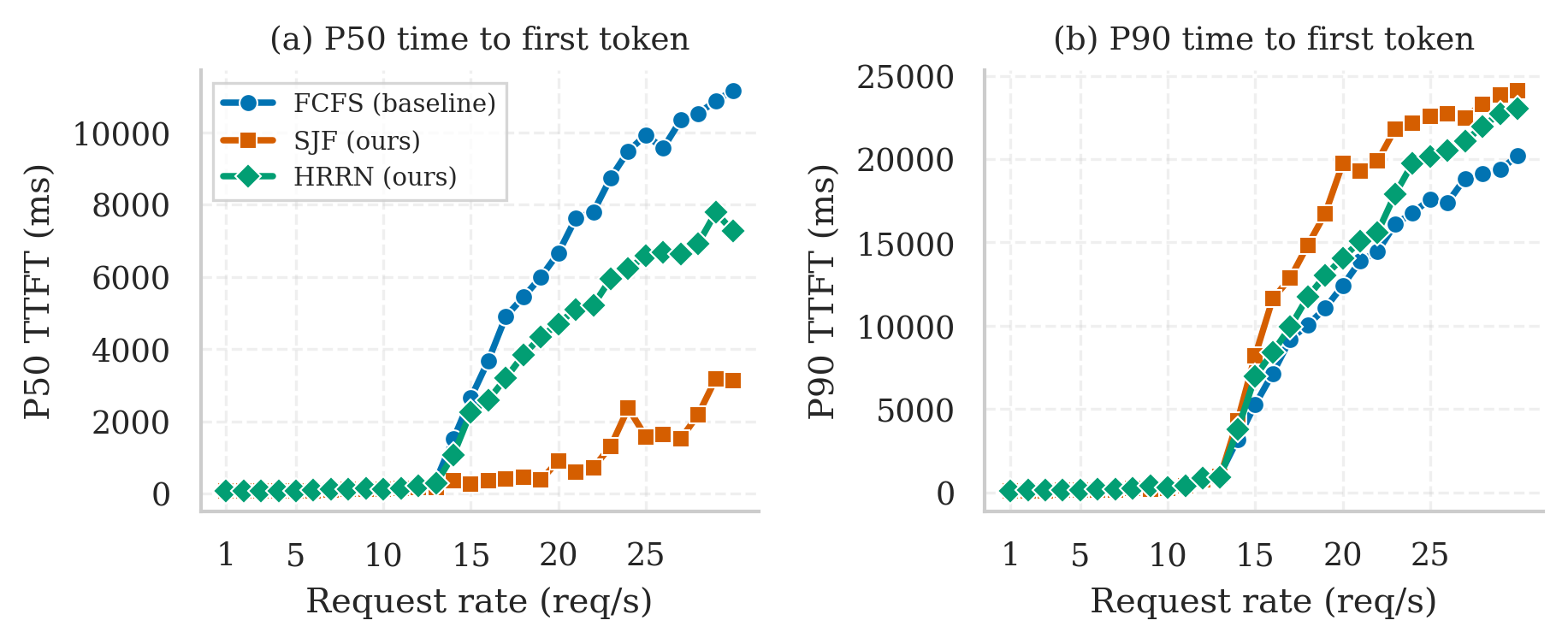}
\caption{Synthetic split TTFT scaling ($1$-$30$ req/s). SJF achieves $-84\%$ $P50$ TTFT at $25$~req/s, confirming that reordering benefits are independent of duration variance.}
\label{fig:ctrl_ttft}
\end{figure}

\begin{figure}[h]
\centering
\includegraphics[width=\textwidth]{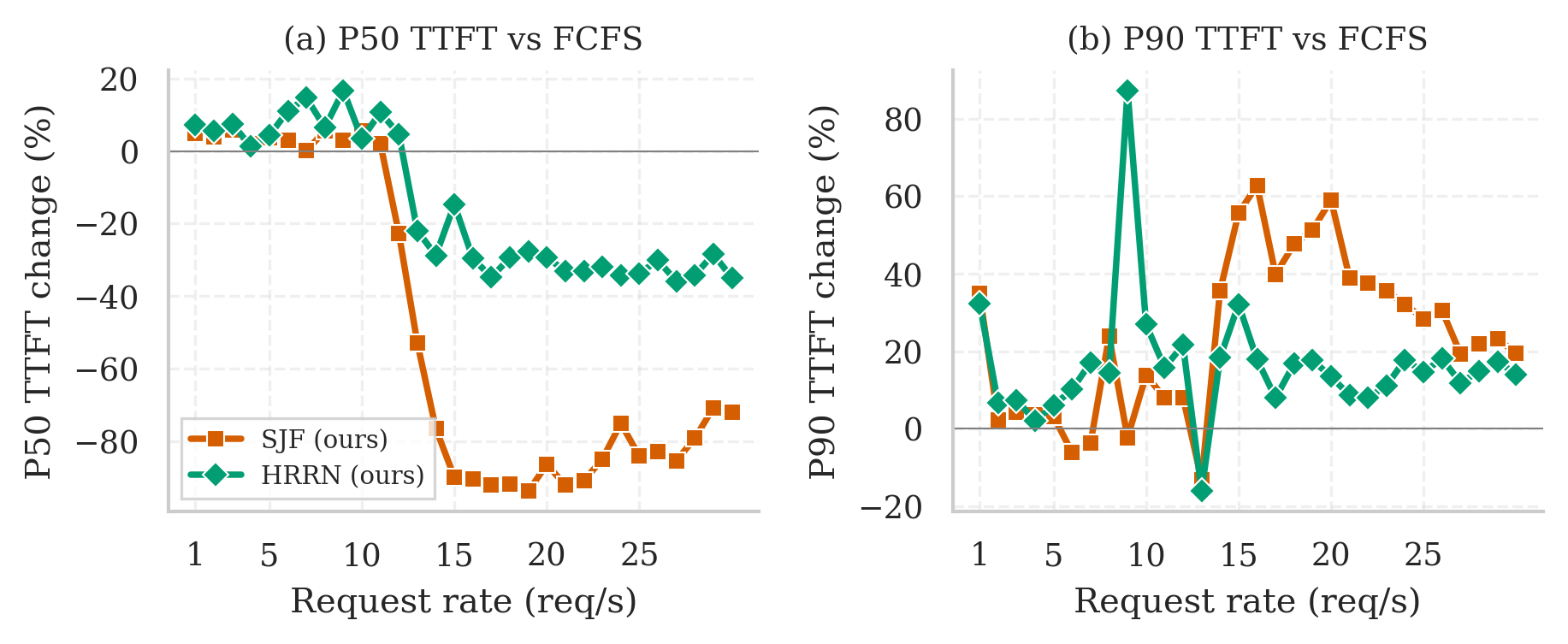}
\caption{Synthetic split: percentage change in TTFT versus FCFS ($1$-$30$~req/s). The pattern closely mirrors LibriSpeech, confirming robustness to duration distribution.}
\label{fig:ctrl_ttft_rel}
\end{figure}

Figure~\ref{fig:ctrl_ttft} shows TTFT on the synthetic split. SJF achieves an $84\%$ $P50$ TTFT reduction at $25$~req/s, confirming that TTFT improvements are driven by queue reordering rather than by exploiting large duration gaps. Percentage changes are shown in Figure~\ref{fig:ctrl_ttft_rel}.

\subsection{Burst Workload Analysis}
\label{app:burst}

The experiments in the main text simulate realistic Poisson arrivals at varying rates. To stress-test scheduling policies under an extreme workload drift scenario, we evaluate a \emph{burst workload} where all $500$ requests arrive simultaneously (request rate $= \infty$). This models sudden traffic spikes common in production~\citep{wang2024burstgpt} and represents the worst-case queuing scenario: the scheduler must immediately triage a deep queue with no inter-arrival gaps.

\begin{figure}[h]
\centering
\includegraphics[width=\textwidth]{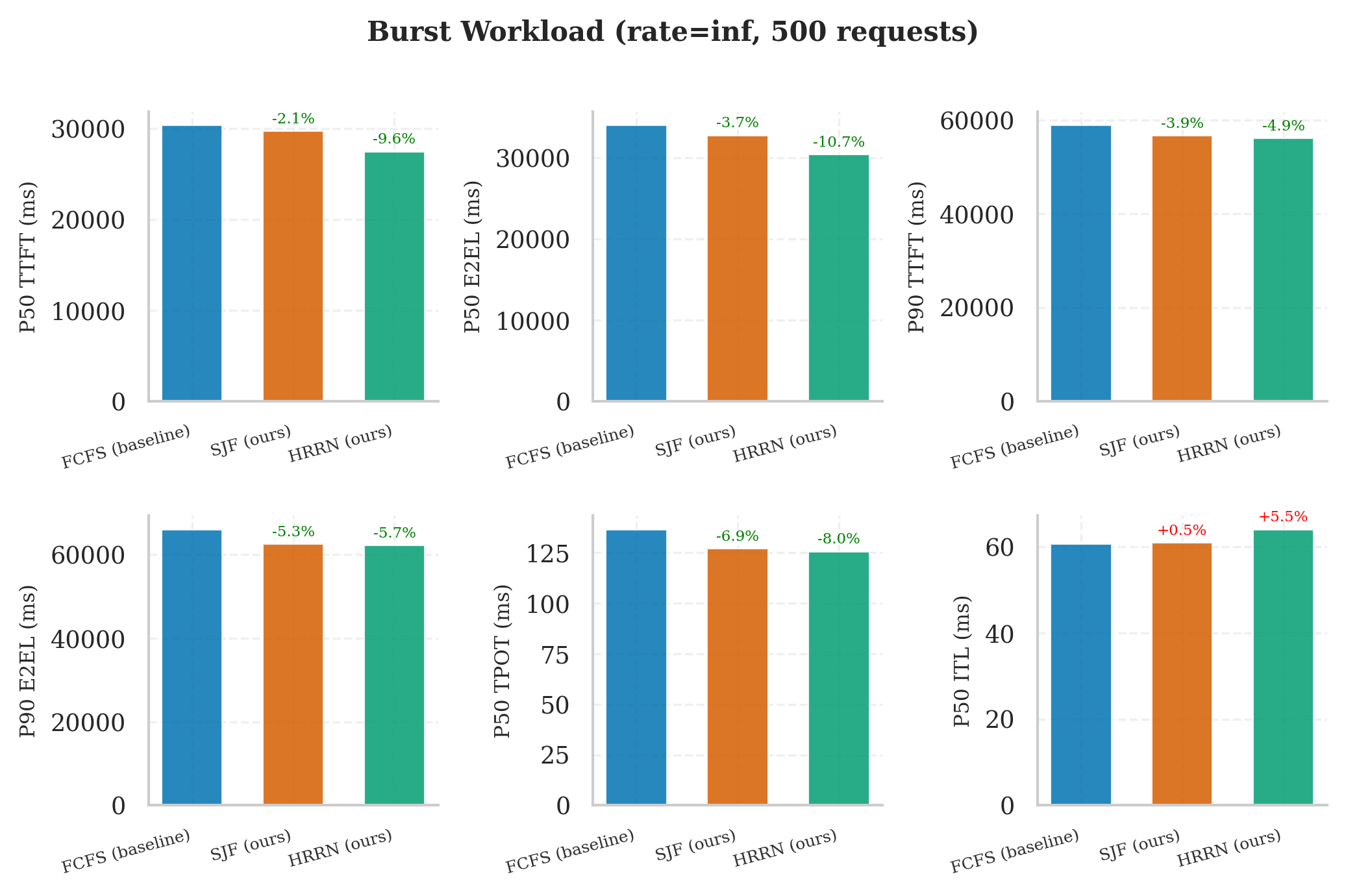}
\caption{Burst workload results (rate$=\infty$, $500$ simultaneous requests). Bar heights show absolute metric values; annotations indicate percentage change versus FCFS. HRRN provides the most consistent improvements across all latency metrics.}
\label{fig:burst}
\end{figure}

\paragraph{Setup.} We run the same whisper-large-v3 model on identical hardware (A100 40\,GB). Each policy processes the full LibriSpeech test-clean split ($500$ utterances) submitted as a single burst. Results are shown in Figure~\ref{fig:burst}.

\paragraph{Results.} Under burst arrival, the differences between policies are smaller than under sustained overload, which is expected since all $500$ requests are queued before any completes. SJF reduces $P50$ TTFT by $-2.1\%$ and $P50$ E2EL by $-3.7\%$ relative to FCFS, while HRRN achieves $-9.6\%$ $P50$ TTFT and $-10.7\%$ $P50$ E2EL. The moderate gains reflect the fact that with a single burst, the queue drains monotonically---there are no new arrivals to continually refresh the short-job advantage that SJF exploits under sustained load.

Time per output token (TPOT) and inter-token latency (ITL) remain largely unaffected ($<8\%$ variation), confirming that scheduling overhead does not impair per-token decoding throughput even under maximum queue depth (500 concurrent requests). HRRN's consistent edge over both FCFS and SJF across all metrics in this setting stems from its ability to balance job priority with accumulated wait time: as the burst progresses, long-waiting requests naturally gain higher response ratios, preventing the extreme starvation that degrades SJF's tail under sustained overload.

\paragraph{Significance.} The burst experiment establishes two practical findings: (1) scheduling benefits persist even under instantaneous load spikes, albeit with smaller magnitude than sustained overload; and (2) HRRN emerges as the preferred policy for unpredictable traffic patterns since it provides consistent improvements without the starvation risk of SJF.

\subsection{Whisper-Medium Model Scaling}
\label{app:whisper_medium}

To assess whether our scheduling findings generalize beyond Whisper-Large-v3 ($1.5$\,B parameters), we repeat the full LibriSpeech evaluation using Whisper-Medium ($769$\,M parameters), roughly half the model size. All other experimental conditions remain identical: the same A100 GPU, identical LibriSpeech test-clean inputs, Poisson arrivals with burstiness~$1.0$, and $500$ completed requests per rate.

\begin{figure}[h]
\centering
\includegraphics[width=\textwidth]{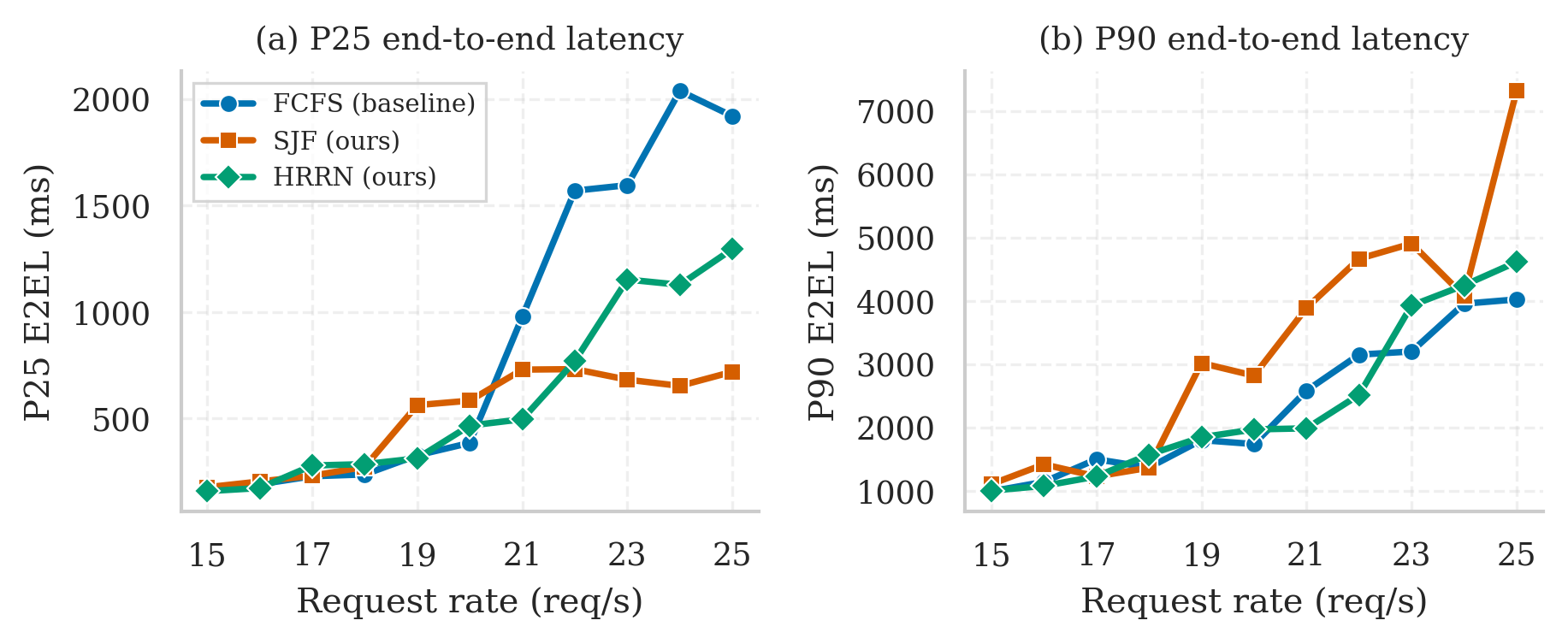}
\caption{Whisper-Medium E2EL scaling ($15$--$25$ req/s). $P25$ latency (a) shows that SJF benefits the bottom quartile of requests across all rates, while $P90$ latency (b) reveals the tail cost at high load, consistent with Large-v3 findings.}
\label{fig:wm_e2el}
\end{figure}

Figure~\ref{fig:wm_e2el} presents $P25$ and $P90$ end-to-end latency for request rates $15$--$25$~req/s, the regime where scheduling effects are most pronounced. Two observations stand out:

\textbf{Higher saturation point.} Whisper-Medium saturates at a higher throughput ($\sim$$21$~req/s versus $\sim$$18$~req/s for Large-v3), which is expected given the smaller model footprint. This shifts the onset of congestion rightward, meaning that scheduling interventions become critical at higher absolute rates.

\textbf{Consistent policy ordering.} Across $15$--$25$~req/s, the relative ranking of policies mirrors Large-v3: SJF delivers the lowest $P25$ E2EL, reflecting its advantage for the majority of (shorter) requests, while HRRN occupies the middle ground. At the $P90$ tail, the familiar starvation trade-off re-emerges at high load, confirming that the median-vs-tail tension is a structural property of priority scheduling rather than a model-specific artifact.

These results are encouraging: the scheduling policies proposed in this work transfer directly to a smaller model without any re-tuning, and the qualitative behavior large P50 gains with a bounded tail cost controlled by HRRN holds across model scales.

\subsection{Implementation Details}
\label{app:implementation}

We integrate SJF and HRRN scheduling into vLLM's v1 engine by modifying five files, totaling $\sim$250 lines of new code. The changes require no modification to the core request serialization format (\texttt{EngineCoreRequest}).

\paragraph{Output token estimation}
At the OpenAI-compatible \texttt{/v1/audio/transcriptions} endpoint, we capture the raw waveform duration $d_i$ before mel spectrogram extraction:

\begin{lstlisting}[style=pycode, caption={Duration-based token estimation at the API endpoint.}]
duration_s = len(audio_data) / sample_rate
estimated_output = max(1, int(duration_s * kappa))
sampling_params.extra_args["estimated_output_tokens"] = estimated_output
\end{lstlisting}

The estimate propagates through \texttt{SamplingParams.extra\_args} to the scheduler without modifying any serialization format.

\paragraph{Request class.}
The \texttt{Request} class reads \texttt{estimated\_output\_tokens} from \texttt{extra\_args} at construction, falling back to \texttt{max\_tokens}:

\begin{lstlisting}[style=pycode, caption={Request class integration (\texttt{v1/request.py}).}]
self.estimated_output_tokens = self.max_tokens
if (sampling_params.extra_args is not None
        and "estimated_output_tokens" in sampling_params.extra_args):
    self.estimated_output_tokens = int(
        sampling_params.extra_args["estimated_output_tokens"])
\end{lstlisting}

\paragraph{Queue implementations.}
\texttt{SJFRequestQueue} uses a min-heap keyed by \texttt{(estimated\_output\_tokens, arrival\_time, request\_id)}, achieving $O(\log n)$ insert/pop. \texttt{HRRNRequestQueue} computes response ratios dynamically at each pop in $O(n)$:

\begin{lstlisting}[style=pycode, caption={SJF and HRRN queue core logic (\texttt{request\_queue.py}).}]
class SJFRequestQueue:
    def _get_priority(self, req):
        return (req.estimated_output_tokens, req.arrival_time, req.request_id)

class HRRNRequestQueue:
    def _response_ratio(self, req, now):
        wait = now - req.arrival_time
        return (wait + req.estimated_output_tokens) / req.estimated_output_tokens
\end{lstlisting}

\paragraph{Preemption logic.}
Under memory pressure, SJF evicts the running request with the \emph{largest} $\hat{n}_i$ (longest job first), while HRRN evicts the request with the \emph{lowest} current $R_i$ (least urgent), keeping preemption consistent with each scheduling objective.

\paragraph{Configuration.}
The scheduling policy is selected via a single CLI flag: \texttt{-{}-scheduling-policy sjf} or \texttt{-{}-scheduling-policy hrrn}. No other configuration changes are required. The full patch is available in the supplementary material.

\end{document}

%% file: math_commands.tex
\usepackage{amsmath,amsfonts,bm}

\def\eqref#1{equation~\ref{#1}}

\def\1{\bm{1}}

\DeclareMathAlphabet{\mathsfit}{\encodingdefault}{\sfdefault}{m}{sl}
\SetMathAlphabet{\mathsfit}{bold}{\encodingdefault}{\sfdefault}{bx}{n}

